\documentclass[graybox]{svmult}

\usepackage{booktabs}
\usepackage{url}
\usepackage{array}
\usepackage{mathptmx}       
\usepackage{helvet}         
\usepackage{courier}        
\usepackage{type1cm}        
\usepackage{makeidx}         
\usepackage{graphicx}        
\usepackage{multicol}        
\usepackage[bottom]{footmisc}
\usepackage[nocompress]{cite}
\usepackage{textcomp}

\makeindex             

\usepackage{amssymb}
\begin{document}

\title*{Utilizing Bidirectional Encoder Representations from Transformers for Answer Selection}
\titlerunning{Utilizing BERT for Answer Selection}
\author{Md Tahmid Rahman Laskar, Enamul Hoque and Jimmy Xiangji Huang}
\institute{Md Tahmid Rahman Laskar \at Department of Electrical Engineering and Computer Science, York University, Toronto, Canada\\ Information Retrieval \& Knowledge Management Research Lab, York University, Toronto, Canada \\ \email{tahmedge@cse.yorku.ca}
\and Enamul Hoque \at School of Information Technology, York University, Toronto, Canada \\ \email{enamulh@yorku.ca}
\and Jimmy Xiangji Huang \at School of Information Technology, York University, Toronto, Canada\\ Information Retrieval \& Knowledge Management Research Lab, York University, Toronto, Canada \\ \email{jhuang@yorku.ca}}

%
%
\maketitle

\abstract{Pre-training a transformer-based model for the language modeling task in a large dataset and then fine-tuning it for downstream tasks has been found very useful in recent years. One major advantage of such pre-trained language models is that they
can effectively absorb the context of each word in a sentence. 
However, for tasks such as the answer selection task, the pre-trained language models have not been extensively used yet. 
To investigate their effectiveness in such tasks, in this paper, we adopt the pre-trained Bidirectional Encoder Representations from Transformer (BERT) language model and fine-tune it on two Question Answering (QA) datasets and three Community Question Answering (CQA) datasets for the answer selection task. We find that fine-tuning the BERT model for the answer selection task is very effective and observe a maximum improvement of 13.1\% in the QA datasets and 18.7\% in the CQA datasets compared to the previous state-of-the-art. 
}

\section{Introduction}
\label{sec:1}
The Answer Selection task is a fundamental problem in the areas of Information Retrieval and Natural Language Processing (NLP)~\cite{yih2013question}. Given a question along with a list of candidate answers, the objective in the answer selection task is to rank the candidate answers based on their relevance with the given question \cite{laskar2019utilizing} (see Table \ref{tab:example}). In such tasks, the relevance between a question and a candidate answer is measured by various sentence similarity modeling techniques~\cite{yih2013question}.

\begin{table}[t!]

\caption{
An example of the Answer Selection task. A question along with a list of candidate answers are given. The sentence in the bold font is the correct answer.}
\small
\begin{center}
\begin{tabular}{p{7cm}}
\hline
\begin{center} \vspace{-3mm} \textbf{Question:} \end{center}
\begin{itemize}     
\item{Who is the winner of the US Open 2019?}     
\end{itemize}
\begin{center} \textbf{List of Candidate Answers:} \end{center}
\begin{itemize}     
\item {{Rafael Nadal has won the French Open 2019.}}     
\item {\textbf{{Rafael Nadal has won the US Open 2019.}}} 
\item {{Roger Federer has won the Australian Open 2018.}} 
\end{itemize}
\begin{center} \textbf{Potential Ranking:}  \end{center}
\begin{itemize}   
\item{\textbf{Rafael Nadal has won the US Open 2019.}} 
\item{Rafael Nadal has won the French Open 2019.} 
\item{Roger Federer has won the Australian Open 2018.}  

\end{itemize}
\\ \hline
\end{tabular}

\end{center}
\label{tab:example} 

\end{table}

In recent years, various sentence similarity models based on the neural network architecture have been utilized to measure the similarity between the question and the candidate answer \cite{chen2018can,chen2018rnn,rao2019bridging}. In such neural models, first, the word embedding  (GloVe\cite{pennington2014glove} or Word2Vec\cite{mikolov2013efficient}) representations of the question and the candidate answer are used as input to the model. Then the vector representations of these sentences produced by the neural model are utilized for the similarity calculation~\cite{chen2018rnn,chen2018can}. However, such word embeddings can only provide a fixed representation of a word and fail to capture its context. Very recently, pre-trained language models have received a lot of attention as they can provide contextual representations of each word in different sentences \cite{peters2018deep,devlin2018bert}. Among the pre-trained language models, fine-tuning the transformer-based \cite{vaswani2017attention} BERT model yields state-of-the-art performance across different NLP tasks \cite{devlin2018bert}. However, the fine-tuned BERT model is not deeply investigated for the answer selection task yet \cite{laskar2019utilizing}.

To be noted that, there are some issues to address regarding fine-tuning a pre-trained model in a new dataset. For instance, the BERT model has been pre-trained in two scenarios: a) when casing information was present, and b) when casing information was absent. Since it is not guaranteed that all datasets will have conventional casing information, it is important to build models that are robust in scenarios when casing information is missing \cite{nercaseduncased}. In addition, it has been observed that neural models which are trained in datasets having conventional casing perform very poorly in the test data for tasks such as named entity recognition \cite{bari2019zero} when the conventional casing is absent \cite{nercaseduncased}. Thus, to address the above issues, in this paper, we fine-tune both the cased and uncased versions of the BERT model for the answer selection task. More concretely, our contributions presented in this paper are the following: 
\begin{itemize}
    \item First, we conduct extensive experiments in five datasets by fine-tuning the BERT model for the answer selection task and observe that the fine-tuned BERT model outperforms all prior work where pre-trained language models were not utilized.  
    \item Second, we show that the cased model of BERT for answer selection is as effective as its uncased counterpart in scenarios when casing information is absent. 
    \item Finally, we conduct ablation study to further investigate the effectiveness of fine-tuning BERT for answer selection. As a secondary contribution, we have made our source codes publicly available here: \url{https://github.com/tahmedge/BERT-for-Answer-Selection}
\end{itemize}

\section{Related Work}
Earlier, various feature engineering-based approaches have been utilized for the answer selection task~\cite{yih2013question,feature2013automatic}. 
However, the feature engineering-based approaches require lots of handcrafted rules and are often error-prone \cite{chen2018can}.
Also, the features which are used in one dataset are not robust in other datasets~\cite{chen2018can}. 

In recent years, several models based on deep neural network have been applied for the answer selection task and they showed impressive performance without requiring any handcrafted features~\cite{kamath2019predicting,chen2018can,chen2018rnn,chen2017enhancing,rao2019bridging,rao2016noise,tay2017learning}. To be noted that, these deep neural network models for answer selection mostly utilized the Recurrent Neural Network (RNN) architecture. However, very recently, models based on the transformer architecture \cite{vaswani2017attention} have outperformed the previously proposed RNN-based models in several NLP tasks \cite{devlin2018bert,liu2019roberta}. Though these transformer-based models utilized the pre-trained BERT architecture \cite{devlin2018bert}, models based on BERT have not been deeply investigated for the answer selection task yet. Moreover, it was found that neural models trained on case sensitive texts performed poorly in scenarios when the conventional casing was missing in the test data \cite{nercaseduncased}. Therefore, to address these issues, we utilize both the cased and uncased versions of the pre-trained BERT model and investigate its generalized effectiveness by conducting extensive experiments in five answer selection datasets. 

\section{Utilizing BERT for Answer Selection}
In this section, we first discuss the transformer encoder \cite{vaswani2017attention} which was utilized in BERT \cite{devlin2018bert}. Then we briefly describe how the BERT model was pre-trained, followed by demonstrating our approach of fine-tuning the pre-trained BERT model for the answer selection task. 
 
\subsection{Transformer Encoder} 
The transformer model has an encoder which reads the text input and a decoder which produces the predicted output of the input text \cite{vaswani2017attention}. The BERT model only utilizes the encoder of transformer \cite{devlin2018bert}. The transformer encoder uses the self-attention mechanism to represent each token in a sentence based on other tokens. This self-attention mechanism works by creating three vectors for each token, which are: a query vector \textnormal{Q}, a key vector \textnormal{K}, and a value vector \textnormal{V}. These three vectors were created by multiplying the embedding vector $\mathbf{x_i}$ with three weight matrices (\textnormal{W\textsubscript{Q}}, \textnormal{W\textsubscript{K}}, \textnormal{W\textsubscript{V}}) respectively. If $\textnormal{d}_k$ is the dimension of the key and query vectors, then the output \textnormal{Z} of self-attention for each word is calculated based on the following:

\begin{equation}
   \textnormal{Z} = softmax\left(\frac{\textnormal{Q} \times \textnormal{K}^{\textnormal{T}}}{\sqrt{\textnormal{d}_k}}\right)\textnormal{V}
\end{equation}

Since the transformer encoder uses multi-head attention mechanism to give attention on different positions, the self attention is computed eight times with eight different weight matrices which provides eight \textnormal{Z} matrices.
Then the eight \textnormal{Z} matrices are concatenated into a single matrix which is later multiplied with an additional weight matrix in order to send the resulting matrix to a feed-forward layer \cite{vaswani2017attention}.

\subsection{Pre-training the BERT Model}
The BERT model adopts the encoder of the transformer architecture \cite{vaswani2017attention}. The encoder of BERT was pre-trained for masked language modeling and the next sentence prediction task on the BooksCorpus (800M words)~\cite{zhu2015aligning} dataset along with the English Wikipedia (2,500M words) \cite{devlin2018bert}. For the masked language modeling task, 15\% tokens in each input sequence are replaced with the special [MASK] token. The model then learns to predict the original value of the masked words based on the context provided by the non-masked words in the input sequence. In the next sentence prediction task, the model receives a pair of sentences as input and attempts to predict if the second sentence in the input pair is a subsequent sentence in the original document. 

\subsection{Fine-tuning BERT for Answer Selection} \label{fine_tuning}

\begin{figure}[t]
\sidecaption[t]
\includegraphics[width=1\linewidth]{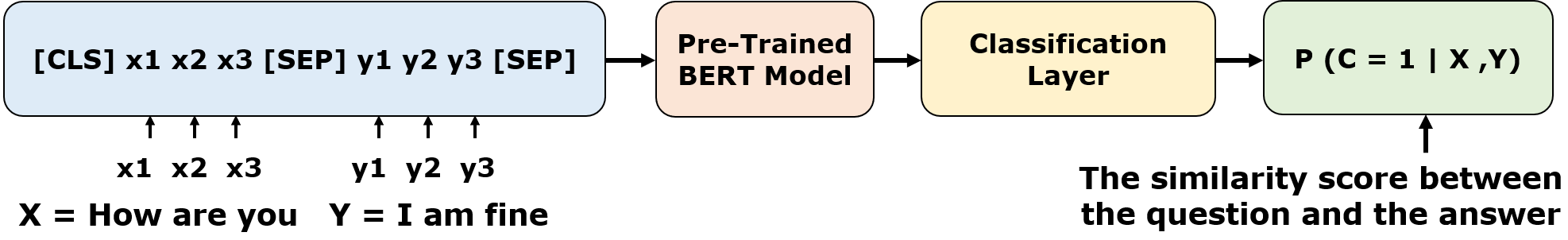}
\caption{BERT Fine Tuning: The question $X$ and the candidate answer $Y$ are combined together as input to the pre-trained BERT model for fine-tuning.}

 \label{fig1}
\end{figure}

Let's assume that we have two sentences $X = x_1, x_2,...,x_m$ and $Y = y_1, y_2,...,y_n$. To input them into the BERT model, they are combined together into a single sequence where a special token $[SEP]$ is added at the end of each sentence. Another special token $[CLS]$ is added at the beginning of the sequence. The fine-tuning process of the BERT model for the answer selection task is shown in Figure \ref{fig1}.

In the fine-tuned BERT model, the representation of the first token ($[CLS]$), which is regarded as the aggregate representation of the sequence, is considered as the output of the classification layer. For fine-tuning, parameters are added to the pre-trained BERT model for the additional classification layer $W$. All the parameters of the pre-trained BERT model along with the additional parameters for the classifier $W$ are fine-tuned jointly to maximize the log-probability of the correct label. The probability of each label $P \in \mathbb{R}^{K}$ (where \textit{K} is the total number of classifier labels) is calculated as follows:
                
                \begin{equation}
                    P = softmax(CW^{T})
                \end{equation}

In the answer selection task, there are two classifier labels (where 1 indicates that the candidate answer is relevant to the question, and 0 indicates the opposite). In the original BERT model \cite{devlin2018bert}, sentence pair classification task was done by determining the correct label. But in this paper, we modify the final layer by following the approach of \cite{laskar2019utilizing} and only consider the predicted score $P_{tr}$ for the similarity label to rank the answers based on their similarities with the question. 
                \begin{equation}
                   P_{tr} = P({C=1} | {X,Y})
                \end{equation}

\section{Experimental Setup}
 In this section, we present the datasets, the training parameters, and the evaluation metrics used in our experiments. To note that all experiments were run using Nvidia V100 with 4 GPUs.

\subsection{Datasets}\label{4.1}

In our experiments, we used five datasets: two of them were Question Answering (QA) datasets whereas rest were Community Question Answering (CQA) datasets. The overall statistics of the datasets are shown in Table~\ref{tab:datasets}. In the following, we give a brief description of each dataset. 

\textbf{TREC-QA:} The TREC-QA dataset is created from the QA track (8-13) of Text REtrieval Conference~\cite{wang2007jeopardy}. 

\textbf{WikiQA:} The WikiQA is an open domain QA
dataset \cite{yang2015wikiqa} in which the answers
were collected from the Wikipedia. 

\textbf{YahooCQA:} The YahooCQA\footnote{\url{https://webscope.sandbox.yahoo.com/catalog.php?datatype=l&did=10}} dataset is a community-based question answering dataset. In this CQA dataset, each question is associated with at most one correct answer and four negative answers \cite{tay2017learning}. 

\textbf{SemEval-2016CQA:} This is also a CQA dataset which is created from the Qatar Living Forums\footnote{\url{https://www.qatarliving.com/forum}}. Each candidate answer is tagged with ``Good", ``Bad" or ``Potentially Useful". We consider ``Good" as positive and other tags as negative \cite{sha2018multi,laskar2020contextualized}.

\textbf{SemEval-2017CQA:} The training and validation data in this CQA dataset is same as SemEval-2016CQA. However, the test sets are different \cite{nakov2017semeval}.

\begin{table*}[t!]
\caption{An overview of the datasets used in our experiments. Here, ``\#" denotes ``number of".}
\centering
\small
\begin{tabular}{c|c|c|c|c|c|c}
\hline

\textbf{Dataset} & \multicolumn{3}{c|}{{\textbf{\# Questions}}}  & \multicolumn{3}{c}{{\textbf{\# Candidate Answers}}}     
\\ \cline{2-7}
& \textbf{Train} & \textbf{Valid} & \textbf{Test} & \textbf{Train} & \textbf{Valid} & \textbf{Test}  
\\ \hline
\textbf{TREC-QA}& 1229 & 82 & 100 & 53417 & 1148 & 1517 \\ \hline
\textbf{WikiQA}& 873 & 126 & 243 & 8672 & 1130 & 2351 \\ \hline
\textbf{YahooCQA}& 50112 & 6289 & 6283 & 253440 & 31680 & 31680 \\
\hline
\textbf{SemEval-2016CQA}& 4879 & 244 & 327 & 36198 & 2440 & 3270 \\
\hline
\textbf{SemEval-2017CQA}& 4879 & 244 & 293 & 36198 & 2440 & 2930 \\ \hline
\end{tabular}
\label{tab:datasets} 
\end{table*}

\subsection{Training Parameters and Evaluation Metrics} \label{TP}

We used both the cased and uncased  models\footnote{\url{https://huggingface.co/transformers/pretrained_models.html}} of BERT\textsubscript{Large} and fine-tuned them for the pairwise sentence classification task~\cite{devlin2018bert}. The parameters of the BERT\textsubscript{Large} model were: number of layers \textbf{L} = 24, hidden size \textbf{H} = 1024, number of self-attention heads \textbf{A} = 16, feed-forward layer size \textbf{d\textsubscript{ff}} = 4096. For implementation, we used the Transformer library of Huggingface\footnote{\url{https://github.com/huggingface/transformers}} \cite{wolf2019transformers}. For training, we used cross entropy loss function to calculate the loss and utilized Adam as the optimizer. We set the batch size to 16 and ran 2 epochs with learning rate being set to $2\times{10}^{-5}$. We selected the model for evaluation which performed the best in the validation set. To evaluate our models, we used the Mean Average Precision (MAP) and the Mean Reciprocal Rank (MRR) as the evaluation metrics.

\section{Results and Analyses}

To evaluate the performance of fine-tuning the BERT model in the answer selection datasets, we compare its performance with various state-of-the-art models \cite{sha2018multi,kamath2019predicting,tay2018hyperbolic,nakov2017semeval}. We also conduct ablation studies to further investigate the effectiveness of fine-tuning. To note that, we pre-processed all datasets into the lower-cased format and evaluated with both the cased and uncased versions of the BERT model.

\newcolumntype{P}[1]{>{\centering\arraybackslash}p{#1}}

\begin{table*}[t!]

\caption{Performance comparisons with the recent progress. Here, `FT' denotes `Fine Tuning'.}

\centering
\small

\begin{center}
\begin{tabular}{c|c|c|c|c|c|c|c|c|c|c}
\hline

\multicolumn{1}{c|}{\textbf{}} &
\multicolumn{4}{c|}{\textbf{QA datasets}} &
\multicolumn{6}{c}{\textbf{CQA datasets}}  \\
\cline{2-11}

\multicolumn{1}{c|}{\textbf{Model}} &
\multicolumn{2}{c|}{\textbf{TREC-QA}} &
\multicolumn{2}{c|}{\textbf{WikiQA}} &
\multicolumn{2}{c|}{\textbf{YahooCQA}} & 
\multicolumn{2}{c|}{\textbf{SemEval\textquotesingle16}} &
\multicolumn{2}{c}{\textbf{SemEval\textquotesingle17}} \\ \cline{2-11}
\centering
 & \textbf{MAP} & \textbf{MRR} &
\textbf{MAP} & \textbf{MRR} &
\textbf{MAP} & \textbf{MRR} &
\textbf{MAP} & \textbf{MRR} &
\textbf{MAP} & \textbf{MRR}  
\\ \hline

\textbf{Kamath et al. \cite{kamath2019predicting}} & 0.852 & 0.891 & - & - & - & - & - & - & - & - \\ \hline

\textbf{Sha et al. \cite{sha2018multi}} & - & - & 0.746 & 0.758 & - & - & 0.801 & 0.872 & - & - 
\\ \hline

\textbf{Tay et al. \cite{tay2018hyperbolic}}& - & - & - & - & - & 0.801 & - & - & - & -
\\ \hline

\textbf{Nakov et al. \cite{nakov2017semeval}}& - & - & - & - & - & - & - & - & 0.884 & 0.928 
\\ \hline

\textbf{BERT\textsubscript{Large (Cased)} FT}  & \textbf{0.934} & \textbf{0.966} & 0.842 & 0.856 & 0.946 & 0.946 & 0.841 & 0.894 & 0.908 & 0.934 \\ \hline
\textbf{BERT\textsubscript{Large (Uncased)} FT} & 0.917 & 0.947 & \textbf{0.843} & \textbf{0.857} & \textbf{0.951} &  \textbf{0.951} & 
 \textbf{0.866} &  \textbf{0.927} &
 \textbf{0.921} &  \textbf{0.963}\\ \hline

\end{tabular}
\end{center}
\label{tab:QACQA} 
\end{table*}

\subsection{Performance Comparisons}

We show the results of our models in Table \ref{tab:QACQA}. We find that in comparison to the prior work in the TREC-QA dataset, the fine-tuned BERT\textsubscript{Large (Cased)} model performs the best and outperforms the previous state-of-the-art \cite{kamath2019predicting} with an improvement of 9.6\% in terms of MAP and an improvement of 8.4\% in terms of MRR. 
However, in the WikiQA dataset, the uncased version performs the best in terms of both MAP and MRR. More specifically, BERT\textsubscript{Large (Uncased)} model improves the performance by 13\% in terms of MAP and 13.1\% in terms of MRR compared to the previous state-of-the-art \cite{sha2018multi} in the WikiQA dataset.

In the CQA datasets, we again observe that both models outperform the prior work. In terms of MRR, we find that the BERT\textsubscript{Large (Uncased)} model outperforms \cite{tay2018hyperbolic}, \cite{sha2018multi}, and \cite{nakov2017semeval} with an improvement of 18.7\%, 6.3\%, and 3.8\% in the YahooCQA, SemEval-2016CQA, and SemEval-2017CQA datasets respectively.

While comparing between the cased model and the uncased model, we find that even though the cased model outperforms the uncased model in the TREC-QA dataset, it fails to outperform the uncased model in rest other datasets. To be noted that, the cased model still provides competitive performance in comparison to the uncased model in all five datasets. 
In order to better analyze the performance of these two models, we conduct significant test. Based on the paired t-test, we find that the performance difference between the two models is \textbf{not statistically significant} ($p$ $\leq$ $.05$). This indicates that the cased version of the fine-tuned BERT model is robust in scenarios when the datasets do not contain any casing information.

\subsection{Ablation Studies}

We perform ablation test to investigate the effectiveness of our approach of fine-tuning the BERT model. For the ablation test, we excluded fine-tuning and only used the feature-based embeddings generated from the pre-trained BERT\textsubscript{Large (Uncased)} model. In our ablation study, we used all five datasets to compare the performance. From the ablation test (see Table \ref{tab:ablation}), we find that removing fine tuning from BERT decreases the performance by 55.8\%, 32.9\%, 54.2\%, 30.3\%, and 24.2\% in terms of MAP in the TREC-QA, WikiQA, YahooCQA, SemEval-2016CQA, and SemEval-2017CQA datasets respectively. The deterioration here without fine-tuning is \textbf{statistically significant} based on paired t-test ($p$ $\leq$ $.05$).

\begin{table*}[t!]

\caption{Performance comparisons based on the Ablation Test. Here, `FT' denotes `Fine Tuning'.}

\centering
\small

\begin{center}
\begin{tabular}{c|c|c|c|c|c|c|c|c|c|c}
\hline

\multicolumn{1}{c|}{\textbf{}} &
\multicolumn{4}{c|}{\textbf{QA datasets}} &
\multicolumn{6}{c}{\textbf{CQA datasets}}  \\
\cline{2-11}
\multicolumn{1}{c|}{\textbf{Model}} &
\multicolumn{2}{c|}{\textbf{TREC-QA}} &
\multicolumn{2}{c|}{\textbf{WikiQA}} &
\multicolumn{2}{c|}{\textbf{YahooCQA}} & 
\multicolumn{2}{c|}{\textbf{SemEval\textquotesingle16}} &
\multicolumn{2}{c}{\textbf{SemEval\textquotesingle17}} \\ \cline{2-11}
\centering
 & \textbf{MAP} & \textbf{MRR} &
\textbf{MAP} & \textbf{MRR} &
\textbf{MAP} & \textbf{MRR} &
\textbf{MAP} & \textbf{MRR} &
\textbf{MAP} & \textbf{MRR}  
\\ \hline

\textbf{BERT\textsubscript{Large (Uncased)} FT} & 0.917 & 0.947 & 0.843 & 0.857 & 0.951 &  0.951 & 
0.866 &  0.927 &
0.921 &  0.963\\ \hline

\textbf{\textit{without} FT} & 0.405 & 0.476 & 0.566 & 0.571 & 0.436 & 0.436 & 0.604  & 0.670  & 0.698 & 0.757 
\\ \hline

\end{tabular}
\end{center}
\label{tab:ablation} 
\end{table*}











\section{Conclusions and Future Work}
In this paper, we adopt the pre-trained BERT model and fine-tune it for the answer selection task in five answer selection datasets. We observe that fine-tuning the BERT model for answer selection is very effective and find that it outperforms all the RNN-based models used previously for such tasks. In addition, we evaluate the effectiveness of the cased version of the BERT model in scenarios when the casing information is not present in the target dataset and demonstrate that the cased model provides almost similar performance compare to the uncased model. We further investigate the effectiveness of fine-tuning the BERT model by conducting ablation studies and observe that fine-tuning significantly improves the performance for the answer selection task.

In the future, we will investigate the performance of different models \cite{laskar2020contextualized} based on the transformer architecture on other tasks, such as information retrieval applications \cite{JH1,JH2,JH3,Jh4,miao2012proximity}, sentiment analysis \cite{JH5,JH6}, learning from imbalanced datasets \cite{JH7}, query-focused abstractive text summarization \cite{laskar2020query,laskar2020wsl}, real-world applications \cite{liu2008modeling}, and automatic chart question answering \cite{kim2020answering}. 

\begin{acknowledgement}
This research is supported by the Natural Sciences \& Engineering Research Council (NSERC) of Canada and an ORF-RE (Ontario Research Fund-Research
Excellence) award in BRAIN Alliance. We acknowledge Compute Canada for providing us with the computing resources and also thank Dr. Qin Chen for helping us with the experiments.
\end{acknowledgement}

\bibliography{ammcs_long}
\bibliographystyle{abbrv}

\end{document}